\documentclass{article}
\usepackage{ijcai19}

\usepackage{times}
\usepackage{soul}
\usepackage{url}
\usepackage[hidelinks]{hyperref}
\usepackage[utf8]{inputenc}
\usepackage[small]{caption}
\usepackage{graphicx}
\usepackage{amsmath}
\usepackage{booktabs}
\usepackage{algorithm}
\usepackage{algorithmic}
\usepackage{multirow}
\usepackage{rotating}
\title{Exploiting Persona Information for Diverse Generation of Conversational Responses}

\author{
    IJCAI19 Anonymous Submission
}

\author{
Haoyu Song$^1$
\and
Wei-Nan Zhang$^{1,2}$
\and
Yiming Cui$^{1,3}$
\and
Dong Wang$^3$\And
Ting Liu$^{1,2}$
\affiliations
$^1$Research Center for Social Computing and Information Retrieval, Harbin Institute of Technology, China\\
$^2$Peng Cheng Laboratory, Shenzhen, China \\
$^3$Joint Laboratory of HIT and iFLYTEK (HFL), iFLYTEK Research, Beijing, China\\
\emails
\{hysong, wnzhang, ymcui, tliu\}@ir.hit.edu.cn,
dongwang4@iflytek.com,
}

\begin{document}

\maketitle

\begin{abstract}
In human conversations, due to their personalities in mind, people can easily carry out and maintain the conversations. Giving conversational context with persona information to a chatbot, how to exploit the information to generate diverse and sustainable conversations is still a non-trivial task. Previous work on persona-based conversational models successfully make use of predefined persona information and have shown great promise in delivering more realistic responses. And they all learn with the assumption that given a source input, there is only one target response. However, in human conversations, there are massive appropriate responses to a given input message. In this paper, we propose a memory-augmented architecture to exploit persona information from context and incorporate a conditional variational autoencoder model together to generate diverse and sustainable conversations. We evaluate the proposed model on a benchmark persona-chat dataset. Both automatic and human evaluations show that our model can deliver more diverse and more engaging persona-based responses than baseline approaches.
\end{abstract}

\section{Introduction}

Building human-like dialogue agents to pass the Turing Test~\cite{Turing1950I} has been a long-term goal of artificial intelligence. Open-domain conversations need to be diversified~\cite{li2015diversity} and sustained~\cite{li2016deepreinforcement}, as the primary goal of an open-domain chatbot is to establish a connection with the user and to accompany the user over a long period of time~\cite{shum2018elizaXiaoIce}. Because of the vast quantity of rational responses for an open-domain conversational message, sequence-to-sequence (Seq2Seq) model~\cite{sutskever2014sequence} has been widely used for conversation modeling~\cite{vinyals2015neural,shang2015neural,serban2016building}. Seq2Seq model enables the incorporation of rich context when mapping between consecutive dialogue turns and is trained by predicting one target response in a given dialogue context using the maximum-likelihood estimation (MLE) objective function.

\begin{table}
\centering
\begin{tabular}{ll}
\hline
\multicolumn{2}{c}{{\bf Chatbot’s Persona }} \\
\hline
1.  I am a soccer player & 4.  Nike cleats are my favorite \\
2.  my number is 42 & 5.  I joined a new team last week \\
3.  I'm a goalie & \\
\hline
\end{tabular}
\end{table}

\begin{table}
\centering
\begin{tabular}{r|cl}  
\hline
\multicolumn{2}{r}{\bf User Input} & What do you do for a living?  \\
\hline
\multirow{6}*{ \rotatebox{90}{\bf Response}} & {Non-Specific}                 & I don’t know what you mean\\
\cline{2-3}
~ & & $r_1$: I am a soccer player, and you?\\
  & {Potential} & $r_2$: Emm... Tell me yours first\\
  & {Candidates} & $r_3$: I am a goalie in the soccer team \ \\
  & & $r_4$: I just joined a new soccer team\\

\hline
\end{tabular}
\caption{Given an input and some persona texts, there exist many non-repetitive responses, e.g., grounded on different persona texts.}
\label{tab:example}
\end{table}

Despite the successful application of Seq2Seq in dialogue modeling, it is still quite impossible for current dialogue agents to pass the Turing Test, while one of the reasons is the lack of a coherent persona~\cite{vinyals2015neural}. Endowing an open-domain chatbot with persona is challenging, yet impactful to deliver more realistic and engaging conversations.
First, presenting a coherent persona is crucial for a chatbot to gain confidence and trust from users~\cite{li2016persona,qian2017assigning}.
Second, grounded on predefined persona, a chatbot can be trained to both ask and answer questions about personal topics, thus making users more engaged~\cite{zhang2018personalizing}. Due to these reasons, there has been a growing interest in modeling persona within the Seq2Seq model.

However, human dialogue has a unique property that many other learning tasks do not: given a dialogue context, there may exist {\bf more than one} valid response\footnote{Denoted as {\it one-to-many} property.}~\cite{zhao2017dialoguecvae,rajendran2018MultipleAnswers}. As illustrated in Table~\ref{tab:example}, the potential candidate responses {\bf cover different aspects} and are {\bf non-repetitive}. Such responses can not only provide more candidates for different dialogue strategies~\cite{yu2016strategy} but also enrich the diversity. Nevertheless, the {\it one-to-many} property is not taken into account in the present persona dialogue models. Although recent Seq2Seq-based persona models were successful, they still suffer from the {\it low diversity} problem~\cite{jiang2018dullsequence} brought by Seq2Seq. They tend to generate repetitive and non-specific responses like {\it “I don’t know”}~\cite{li2015diversity,li2016deepreinforcement}, regardless of the conversational context.

In this paper, we follow the definition of persona in Zhang {\it et al.}~\shortcite{zhang2018personalizing} (as exemplified in Table~\ref{tab:example}), and work on the explicit textual persona. This is because the unstructured persona data is easier to collect and more authentic\footnote{ For example, 80\% of posts on Twitter are about personal emotional state, thoughts or activities~\cite{naaman2010messagecontent}.}. 
We propose a memory-augmented architecture to exploit the explicit persona texts and incorporate a latent variable that can capture response variations under the conditional variational autoencoder model, rather than learn to predict only one response using MLE.
To generate diverse persona-based responses, our model draws samples from the latent variable and combines two different decoding strategies to generate words or copy words directly from persona texts. We evaluate our model on the ConvAI2 persona-chat dataset and yield promising results in (i) incorporating persona information into responses and (ii) generating diverse and engaging responses.
In summary, the contributions of this paper are three-fold:
\\{\bf First}, we propose to address the {\it one-to-many} property in end-to-end model. To the best of our knowledge, this is the first work that jointly models persona text and response diversity.
\\{\bf Second}, we present a novel end-to-end model that incorporates persona texts into diverse responses (Section~\ref{tab:proposed_model}).
\\{\bf Third}, experimental results show that our model can generate persona-based responses as well as deliver more diverse and more engaging responses than baselines (Section~\ref{tab:experiment}).

\section{Related Work}

This work is related to the recent advancements in persona-based dialogue models and the efforts of improving diversity in dialogue generation.

In recent years, the dialogue research community has shown a growing interest in the modeling of personality. Li {\it et al.}~\shortcite{li2016persona} first incorporated implicit persona in dialogue generation using user embeddings, which projects each user into a dense vector. Kottur {\it et al.}~\shortcite{kottur2017exploringpersona} proposed a neural dialogue model which conditioned on both speakers and context history. However, these two models rely heavily on speaker-tagged dialogue data, which is more expensive and sparser.  To present a coherent personality, Qian {\it et al.}~\shortcite{qian2017assigning} defined several profile key-value pairs, including name, gender, age, location, etc., and explicitly expressed a profile value in the response. Recent works brought new persona models as well as high-quality data. Zhang {\it et al.}~\shortcite{zhang2018personalizing} contributed a persona-chat dataset, and they further proposed two generative models, {\it persona-Seq2Seq} and {\it Generative Profile Memory Network}, to incorporate persona information into responses. Yavuz {\it et al.}~\shortcite{deepcopy} explored the use of copy mechanism in persona-based dialogue models. Our work follows the line initiated by Zhang {\it et al.}~\shortcite{zhang2018personalizing} but makes a step further by generating diverse persona-based responses.

On the other hand, large-scale dialogue generation still suffers from the tendency of generating dull and generic responses. The efforts to tackle the low diversity problem fall into two major categories. The first focuses on the objective function of the dialogue generation model, and they argue that the MLE objective is unable to approximate the real-world goal of the conversation~\cite{li2015diversity,li2016deepreinforcement}. The other line of research tries to augmenting the encoder of the generative model with richer information~\cite{xing2016topic,zhao2017dialoguecvae}. 
As one of these improvements, Zhao {\it et al.}~\shortcite{zhao2017dialoguecvae} proposed a dialogue model adapted from conditional variational autoencoders~\cite{yan2016cvae}. The dialogue CVAE learns a latent variable that captures discourse-level variations and generates diverse responses by drawing samples from the learned distribution. 

We also borrow ideas from Sukhbaatar {\it et al.}~\shortcite{sukhbaatar2015end2endmn} in the modeling of external memory. The proposed Persona-CVAE can be seen as an extension of dialogue CVAE, with several main differences: (i) the combination with external persona memory, (ii) the decoding strategies, and (iii) the ability to deliver persona-based responses.
\label{tab:cvaemodel}

\begin{figure}[ht]
  \centering
  \includegraphics[width=1.0\linewidth]{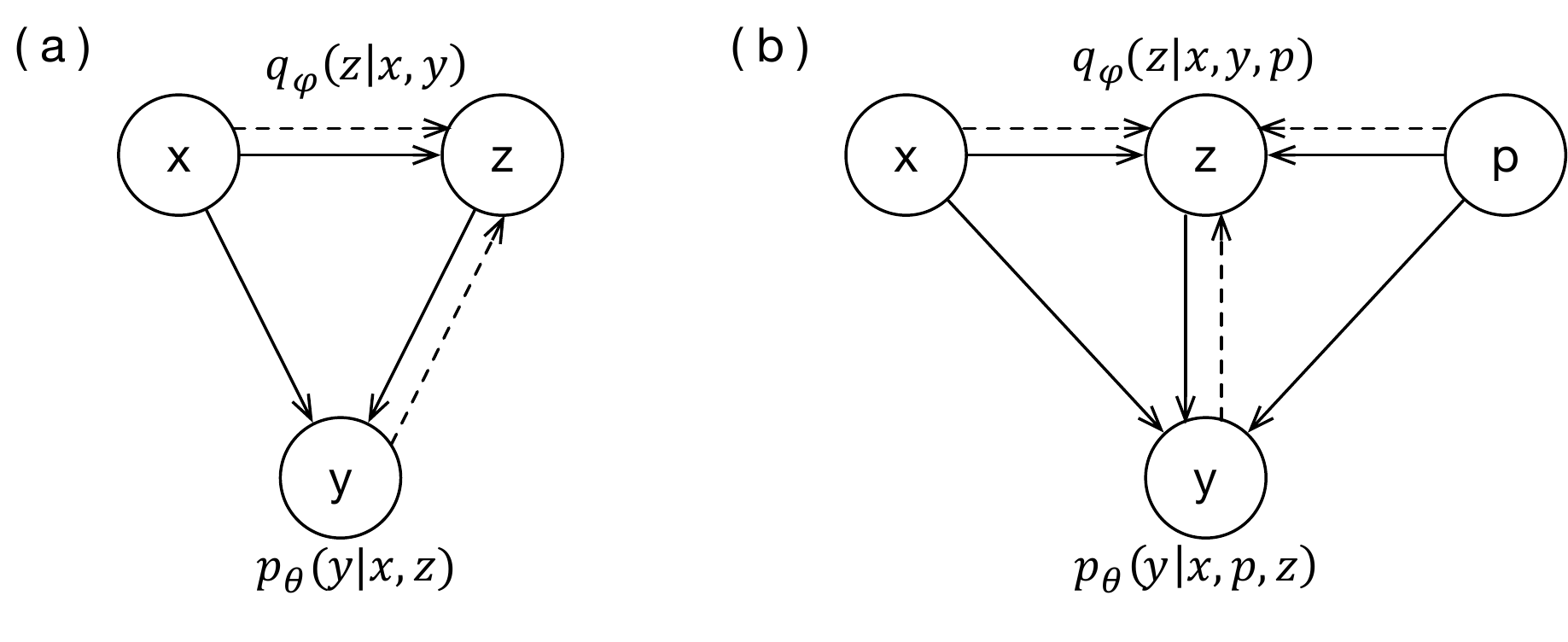}
  \caption{Illustration of the conditional graphical models: (a) dialogue CVAE by Zhao and (b) the proposed Persona-CVAE. Solid lines denote the generative model with parameter $\theta$, dashed lines denote the variational approximation with parameter $\varphi$. {\it x}, {\it y}, {\it z} and {\it p} denote input, response, latent variable and persona respectively.} 
  \label{fig:cvae_graphmodel}
\end{figure}

\section{Persona-CVAE}
\label{tab:proposed_model}

In this section, we describe the components of the proposed Persona-CVAE model in detail.
As illustrated in Figure~\ref{fig:cvae_graphmodel} (b), the conditional graphical model shows the core idea of proposed Persona-CVAE. We assume that the generation of $z$ depends on $x$, $y$ and $p$. y relies on x, p and z. The main objective is to learn the conditional probability $q_\varphi(z|x,p)$ and $p_\theta(y|x,p,z)$, with parameters $\varphi$ and $\theta$. The variational parameters $\varphi$ are learned jointly with the generative model parameters $\theta$, as described in Section \ref{sec:cvae}.

\begin{figure*}[ht]
  \centering
  \includegraphics[width=1.0\linewidth]{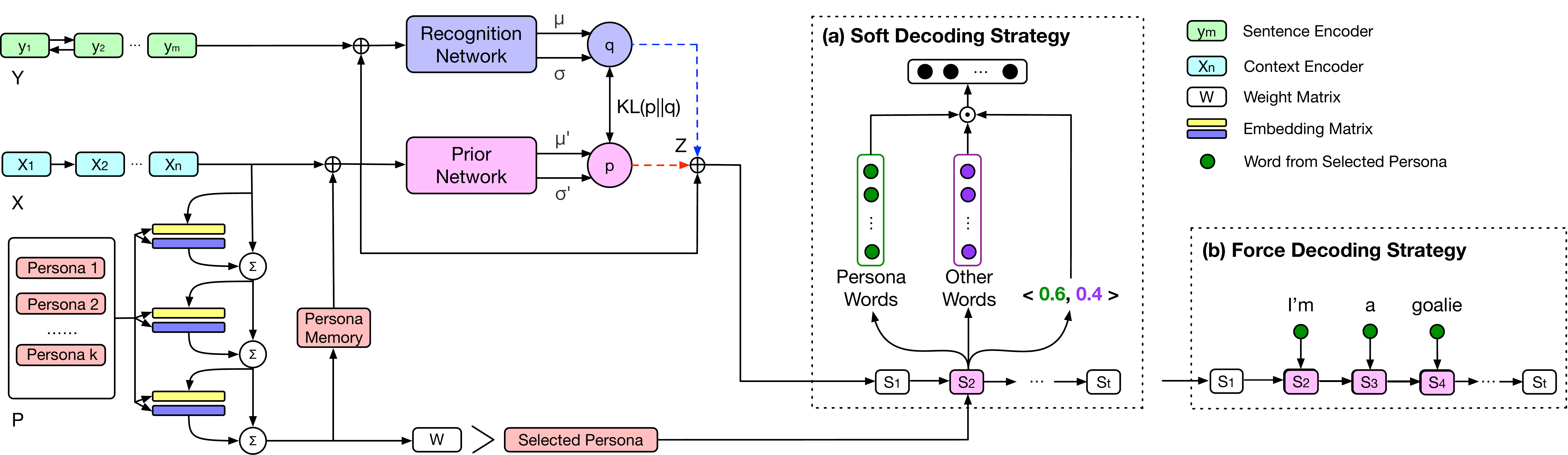}
  \caption{The neural network architecture of the proposed Persona-CVAE model. The persona memory comes from the multi-hops attention over persona texts and dialogue context. The dashed blue connection only appears in the training process, and the dashed red connection only appears in the generation process. The $\bigoplus$ denotes the concatenation of input vectors.}
  \label{fig:main_model}
\end{figure*}

\subsection{Task Definition}
The task can be formally defined as: given an input message $X$ and a set of persona texts $P={\{P_1,P_2,...,P_n\}}$, the goal is to generate diverse responses $\hat Y=\{\hat Y_1,\hat Y_2,...,\hat Y_m\}$, based on both input message and persona texts.

The {\bf diverse} in responses means that the utterances in $\hat Y$ are non-repetitive, cover different aspects and differ in words.

\subsection{Persona based CVAE}

The proposed model uses four random variables to represent each dyadic conversation: the dialogue context $X$, the target response $Y$, the set of persona texts $P$ and a latent variable $z$. $z$ is used to capture the latent distribution over the valid responses. Further, $X$ is composed of both input messages and responses in dialogue history and can be denoted as $X=\{X_1,X_2,...,X_n\}$, where $X_i=x_{i,1}x_{i,2}...x_{i,k}$ ($x_{i,j}$ is a single word). Similarly, $P$ consists of several unstructured persona texts: $P=\{P_1,P_2,...,P_k\}$, where $P_i=p_{i,1}p_{i,2}...p_{i,l}$.

Figure \ref{fig:main_model} demonstrates an overview of our model. 
The sentence encoder is a bidirectional recurrent neural network (BRNN)~\cite{schuster1997brnn} and is used to encode a single sentence (e.g., the target response). We concatenate the last hidden state of both forward and backward RNN to capture the semantic information from both sides, formulized as $h=[\overrightarrow{h_T},\overleftarrow{h_T}]$, where $T$ is the sequence length. 
For the context encoder, we use a hierarchical encoding strategy. First, each sentence $X_i$ in context $X$ is encoded by the sentence encoder to get a latent representation $h_{sent_i}$. Then a single layer forward RNN is used to encode the sentence representations into a final state $h_{context}$. 

Persona differs from dialogue context in two main aspects: (i) it keeps unchanged throughout the conversation, and (ii) it is only the unilateral information (here is for the chatbot), while two sides share the same dialogue context. Therefore, we conjecture that modeling persona differently from dialogue context, with a memory augmented architecture, will be beneficial for the model's performance. The details will be discussed in Section~\ref{sec:persona_memory}.

Then we define the conditional distribution over the above random variables. As mentioned, we define the generation process as a conditional distribution $p(y,z|x,p) = p(y|x,p,z)p(z|x,p)$ and our objective is to approximate $p(y|x,p,z)$ and $p(z|x,p)$ with deep neural networks. Following the definition in Zhao {\it et al.}~\shortcite{zhao2017dialoguecvae}, we refer to $p(y|x,p,z)$ as the {\it response decoder} and $p_{\theta}(z|x,p)$ as the {\it prior network}. Further, to approximate the true posterior distribution, we refer to $q_{\varphi}(z|x,y,p)$ as the {\it recognition network}.

CVAE is trained to maximize the conditional log-likelihood, but this involves an intractable marginalization over the latent variable. Previous works~\cite{sohn2015learning,yan2016cvae} have shown that CVAE can be efficiently trained with the {\it Stochastic Gradient Variational Bayes} (SGVB) \cite{kingma2013vae} by maximizing the {\it variational lower bound} of the conditional log-likelihood. As proposed in Zhao {\it et al.}~\shortcite{zhao2017dialoguecvae}, we also assume the latent variable $z$ follows multivariate Gaussian distribution with a diagonal covariance matrix. And the {\it variational lower bound} of Persona-CVAE can be written as:
\begin{align}
\label{equo:vlb}
    \mathcal{L}(\theta,\varphi;x,y,p,z) &= -KL(p_{\theta}(z|x,p)||q_{\varphi}(z|x,y,p)) \nonumber \\
    &+ E_{q_{\varphi}(z|x,y,p)}(\log p(y|x,p,z))\\
    &\leq \log p(y|x,p) \nonumber
\end{align}
Since we assume latent variable $z$ follows isotropic Gaussian distribution, the recognition network $q_{\varphi}(z|x,y,p) \sim \mathcal{N}(\mu,\sigma^{2}{\bf I})$ and the prior network $p_{\theta}(z|x,p) \sim \mathcal{N}(\mu',\sigma'^{2}{\bf I})$. We sample $z$ either from $\mathcal{N}(\mu,\sigma^{2}{\bf I})$ in training or $\mathcal{N}(\mu',\sigma'^{2}{\bf I})$ in testing. To make the sampling operation differentiable, we use the {\it reparametrization trick}~\cite{kingma2013vae} and we have:
\begin{align}
\left[ \begin{array}{c}
\mu\\
\sigma^{2}
\end{array}
\right ]&=
W_{recog}\left[ \begin{array}{c}
x \\
y \\
p
\end{array}
\right ]+b_{recog}\\
\left[ \begin{array}{c}
\mu'\\
\sigma'^{2}
\end{array}
\right ]&=
W_{piror}\left[ \begin{array}{c}
x \\
p
\end{array}
\right ]+b_{piror}
\end{align}
We project the concatenation of $x$,$p$ and $z$ to a vector, to serve as the initial state of {\it response decoder}.
Then the {\it response decoder} predicts words under the Decoding Strategies (Section~\ref{sec:decoding_stra}).

\label{sec:cvae}

\subsection{Persona Memory}
The persona memory module has two roles: first is to encode the persona texts ${P=\{p_1,p_2,...,p_k\}}$ into a dense vector (persona memory) and sencond is to select a persona $p_i$ to be expressed in the response. Notice that persona is not always needed in the response, and thus model selects persona from ${\{p_1,p_2,...,p_k\}}\cup\{\mbox{None}\}$.

The {\bf first role} of the persona memory module is defined by multi-hops attention over persona texts and dialogue context. We convert the persona texts into memory vectors ${\{m_1,m_2,...,m_k\}}$ of dimension $d$, computed by embedding each $p_i$ in a continuous space using an embedding matrix $A$ of size $d \times V$. We use the embedded dialogue context $h_{context}$ (with the same dimensions $d$) as an input. We compute the match between context vector $h_{context}$ and each memory vector $m_i$ in the embedding space by:
\begin{align}
    prob_i=\mathbf{softmax}(h_{context}^\mathrm{T}m_i)
\label{equo:prob}
\end{align}
where $\mathbf{softmax}(x_i)=e^{x_i} / \sum_je^{x_j}$. Meanwhile, each persona $p_i$ has an output vector $c_i$ (by another embedding matrix $C$). A single hop output $o$ is computed by a sum over the output vector $c_i$, weighted by the probability from 
equation~(\ref{equo:prob}):
\begin{align}
    o=\sum_iprob_ic_i
\end{align}
And $k$-hops attention are stacked in the following way:
\begin{align}
    u^{k}=u^{k-1} + o^{k-1}
\end{align}
where $u^0=h_{context}$. For the embedding matrix, we use an adjacent strategy, i.e. $A^{k+1}=C^{k}$. Finally, the $u^3$ is used as our persona memory{\footnote{In our experiments, $k=3$ performs better than $k=1$ or $2$ , but there is no significant performance change when $k=4$ or $5$.}}.

The {\bf second role} of the persona memory module is to decide which persona should be expressed in the generated response. This is implemented as:
\begin{align}
    \alpha_i = \mathbf{softmax}(\mathbf{W_p}[u^3,z]) = \mathbf{MLP}([u^3,z])
\end{align}
where $\mathbf{W_p}$ is a weight matrix, and $u^3$ is the persona memory. $z$ is the sampled latent variable, as described in Section~\ref{sec:cvae}. The latent variable captures some information that is not presented in the dialogue context and is beneficial for selecting the optimal persona. As aforementioned, in some cases, there is no need to incorporate persona information, so we need an MLP here to make the decision. The model selects a persona with the maximal probability, i.e. $\hat{p}=p_j$, where $j={\mathbf {argmax}_i}(\alpha_i)$. Then the words from selected persona are further used in {\it response decoder}.

\label{sec:persona_memory}

\subsection{Decoding Strategy}
We propose two decoding strategies (general purpose and special cases) to better express persona information.

\subsubsection{Soft Decoding Strategy (SDS)}
The {\it soft decoding strategy} assumes that at each decoding step $t$, there is a distribution $\alpha^t$ over the two types {\it \{persona, other\}}. Similarly, according to the selected persona words, the entire vocabulary set is also divided into two parts, i.e. {\it \{persona words, other words\}}.

At time step $t$, the {\it response decoder} first estimates word generation probabilities over the two vocabulary sets independently, denoted as ${\mathcal{P}_{per}^t}$ and ${\mathcal{P}_{other}^t}$:
\begin{align}
    {\mathcal P}_{per}^t &= {\mathbf{softmax}_{per}}(w_{dec}s_t+b_{dec}) \\
    {\mathcal P}_{other}^t &= {\mathbf{softmax}_{other}}(w_{dec}s_t+b_{dec})
\end{align}
and then computes the type distributions:
\begin{align}
\left[ \begin{array}{c}
\alpha_{per}^t \\
\alpha_{other}^t
\end{array}
\right ]=
{\mathbf{softmax}}(w_{sds}\left[ \begin{array}{c}
s_t \\
u^3
\end{array}
\right ] + b_{sds})
\end{align}
where $s_t$ is the hidden state of decoder RNN cell. The final probability of generating a word is a mixture of type-specific generation distributions where the coefficients are type probabilities:
\begin{align}
    {\mathcal P}(y_t|Y_{<t},X,P)=\alpha_{per}^t{\mathcal{P}_{per}^t} \cup \alpha_{other}^t{\mathcal{P}_{other}^t}
\end{align}

\subsubsection{Force Decoding Strategy (FDS)}
We propose {\it force decoding strategy} from the observation that persona texts can directly serve as a valid response under some circumstances. In the decoding process, if the last words of the partially decoded sequence are the same as the first words of selected persona, the input words to decoder RNN cell will come from the selected persona words successively in next few steps:
\begin{align}
  {h}_{t+1} = \mathbf{RNN}_{dec}({h}_{t}, \mathbf{embedding}(p_i))
\end{align}
where $p_i$ is from the selected persona. At time step $t$, $p_i$ is also the output word. The decoder uses FDS only once in a decoding process. Though this strategy is simple, we found it works well in some situations.
\label{sec:decoding_stra}

\subsection{Training and Optimization}
The Persona-CVAE is trained to maximizing the {\it variational lower bound} of the conditional log-likelihood in Formula~(\ref{equo:vlb}). Persona memory module and type distribution are trained by two standard cross-entropy loss function.

As addressed in Bowman {\it et al.}~\shortcite{Bowman2015cvaekl}, the {\it KL annealing} trick is necessary for the training of RNN-based CVAE. Besides, another essential technique for the model to work is the {\it bag-of-word loss}~\cite{zhao2017dialoguecvae}. Finally, the Persona-CVAE model is trained with the sum of these losses and optimized through backpropagation.

\section{Experiment}
\label{tab:experiment}
\subsection{Persona-Chat Dataset and Preparation}
\label{tab:dataPreparation}
We perform experiments on the recently released ConvAI2 benchmark dataset, which is an extended version (with a new test set) of persona-chat dataset~\cite{zhang2018personalizing}. The conversations are obtained from crowdworkers who were randomly paired and asked to act the part of a given persona.

This dataset contains 164,356 utterances in over 10,981 dialogues and has a set of 1,155 personas, each consisting of at least four profile texts. The testing set contains 1,016 dialogues and 200 never seen before personas. We set aside 800 dialogues together with its profile texts from the training set for validation. The final data have 9,181/800/1,016 dialogues for train/validate/test.

To train the persona memory module, we label each utterance with its corresponding persona.We first compute word inverse document frequency: $idf_i=1/(1+\log(1+tf_i))$, where {\it $tf_i$} is from the GloVe index via Zipf’s law~\cite{zhang2018personalizing}. If the utterance has the highest {\it tf-idf} similarity with a profile sentence and the similarity is higher than a threshold, then we label the utterance with this profile sentence. Otherwise, the utterance's persona label is none.
We also label the position that shares a word with profile sentence to learn the type distribution.

\begin{table*}
\centering
\begin{tabular}{l|ccc|ccc|ccc}
\hline
\multirow{2}*{\bf Methods}& &{\bf N = 1}& & &{\bf N = 5}& & &{\bf N = 10}& \\
\cline{2-10}
&{\bf Dtinct-1}&{\bf Dtinct-2}&{\bf P. Cover}&{\bf Dtinct-1}&{\bf Dtinct-2}&{\bf P. Cover}&{\bf Dtinct-1}&{\bf Dtinct-2}&{\bf P. Cover}\\
\hline
Seq2Seq& .0125 & .0464 & .0026 & .0031 & .0142 & .0057 & .0018 & .0089 & .0071 \\
CVAE& .0366 & .2080 & .0021 & .0090 & .0875 & .0029 & .0050 & .0663 & .0048 \\
\hline
Per.-Seq2Seq& .0159 & .0745 & .0091 & .0036 & .0213 & .0217 & .0021 & .0139 & .0193 \\
GPMN& .0179 & .0738 & .0080 & .0045 & .0195 & .0184 & .0027 & .0103 & .0178 \\
Oracle. Copy& .0203 & .0830 & {\bf .0181} & .0050 & .0276 & .0273 & .0031 & .0189 & .0247 \\
\hline
Per.-CVAE& {\bf .0383$^{\star}$} & {\bf .2088} & .0167 & {\bf .0120$^{\star}$} & {\bf .1037$^{\star}$} & {\bf .0410$^{\star}$} & {\bf .0075$^{\star}$} & {\bf .0779$^{\star}$} & {\bf .0395$^{\star}$}\\
\hline
\end{tabular}
\caption{
Automatic evaluation results in diversity ({\it Dtinct-1\&2}) and persona integration ({\it P. Cover}). $N$ is the number of generated responses in each turn. 
Numbers with the $\star$ mean that the result is statistically significant over all baselines (2-tailed t-test, p-value$<$0.01). 
}
\label{tab:auto_distinct}
\end{table*}

\subsection{Baselines}

We compared the proposed Persona-CVAE model with five state-of-the-art generative baseline models. The five models fall into two categories, i.e., persona-free model and persona-based model:\\
{\bf Seq2Seq}: a general persona-free model with context attention mechanism~\cite{shang2015neural}. \\
{\bf Dialogue CVAE}: a persona-free model that uses a latent variable to learn a distribution over potential conversational intents and improves the discourse-level diversity of responses~\cite{zhao2017dialoguecvae}. \\
{\bf Persona-Seq2Seq}: a persona-based model that prepends persona texts to the input sequence $x$, i.e., $x=\forall{p}\in{P} ||x$,where $||$ denotes concatenation~\cite{zhang2018personalizing}.\\
{\bf Generative Profile Memory Network (GPMN)}: a persona-based generative model that encodes each profile sentence as individual memory representations in a memory network~\cite{zhang2018personalizing}.\\
{\bf Oracle Seq2Seq+Copy (Oracle. Copy)}: a persona-based model with copy mechanism. This model is similar to Persona-Seq2Seq, but it selects the persona text whose tokens have the highest unigram {\it tf-idf} similarity to the {\bf ground truth} response (the {\it Oracle}). It achieves the best results in a variety of copy-based persona models~\cite{deepcopy}. The aim of this experiment is to have a better comparison with the copy-based persona models, although using the ground truth for persona selection is not fair.

Among these baselines, we fine-tuned Seq2Seq, Dialogue CVAE and Oracle Seq2Seq+Copy on the ConvAI2 persona-chat dataset, while the other two methods are using the latest released models from Zhang {\it et al.}~\shortcite{zhang2018personalizing} in ParlAI.
Moreover, to make different models comparable, we generate $N$ responses from the Seq2Seq based models by sampling from the softmax. For the latent variable based models, we sample $N$ times from the latent $z$ to generate responses.

\subsection{Experimental Settings}
In our experiments\footnote{Code available at: \url{https://github.com/vsharecodes/percvae} .}, 
the RNN is two-layer GRU with a 500-dimensional hidden state. The dimension of word embedding is set to 300, and thus the persona memory size is also 300. The vocabulary size is limited to 20,000. The latent variable size is set to 100. KL annealing steps are set to 10,000. We train the model with a minibatch size of 32 and use Adam optimizer with an initial learning rate of 0.001. All parameters are initialized by sampling from a uniform distribution.

\subsection{Automatic Evaluation}

Automatically evaluating an open-domain generative dialogue model is still an unsolved research challenge. Although metrics such as BLEU and perplexity have been used for dialogue quality evaluation~\cite{vinyals2015neural,serban2016building}, recent work has found that BLEU shows very weak correlation with human judgment ~\cite{Liu2016Evaluate}. Further, perplexity is not computed on a per-response basis. Since the goal of the proposed model is not to predict only one target response, but rather diverse persona-based responses, we do not employ BLEU or perplexity for evaluation. Following the {\it one-to-many} property, we use two metrics to evaluate how well our model enhancing diversity and incorporating persona information:\\
{\bf Distinct-K (Dtinct-K)}: this metric calculates the number of distinct k-grams in generated responses and is scaled by the total number of generated tokens to avoid favoring long responses~\cite{li2015diversity}. The Distinct-1 and Distinct-2 are thus the token ratios for unigrams and bigrams. This metric is an indicator of word-level diversity for generated responses.
\\
{\bf Persona Coverage (P. Cover)}: we propose this metric to evaluate how well persona information is expressed. We assume that there are $M$ predefined profile sentences $\{p_1,p_2,...,p_M\}$ and model generates $N$ hypothesis responses $\{\hat{y}_1,\hat{y}_2,...,\hat{y}_N\}$. The response-level persona coverage is defined as:
\begin{align}
    {\mathcal C}_{per} &= \frac{\sum_i^N{\mathbf{max}_{j\in{[1,M]}}\mathcal{S}(\hat{y}_i,p_j)}}{N}
\end{align}
where $\mathcal{S}(\hat{y}_i,p_j)$ is the counts of shared words weighted by words' {\it idf} (as described in Section~\ref{tab:dataPreparation}). Assume that the words' set shared by $\hat{y}_i$ and $p_j$ are $W$ and the {\it inverse document frequency} for $w_k\in{W}$ is $f_k$, then we have:
\begin{align}
    \mathcal{S}(\hat{y}_i,p_j) &= \frac{\sum_{w_k\in{W}}{({f_k})}}{|W|}
\end{align}
and $|W|$ denotes the number of words in $W$.

The final score is averaged over the entire test dataset, and we report the performance in Table~\ref{tab:auto_distinct}.
As can be seen, even if $N$=1, our model still outperforms all the baselines in diversity but is inferior to the Oracle model in the {\it Persona Coverage}. When $N$=5 and 10, the proposed model obtains the best performance in all metrics compared with all the baseline models. Due to the number of persona texts in each dialogue (generally 4 or 5), generating too many responses leads to a small decline in the {\it Persona Coverage}.

The results show that under the different number of $N$, our model always generates diverse responses as well as effectively incorporates persona information into multiple responses.
To better evaluate the diversity of generated responses, the following experiments are all based on $N$=5.

\subsection{Human Evaluation}
We also explore two settings for human evaluation. In both settings, we present persona texts, input message, as well as $N$ generated responses.

In the first setting, we employ judges to evaluate a random sample of 100 items per model according to three metrics, based on a 1/0 scoring schema, similar as Qian {\it et al.}~\shortcite{qian2017assigning}:\\
{\bf Engagingness (Engage.)}: a response is engaging only when it is appropriate, interesting and easy to answer. This is the overall score of $N$ generated responses.\\
{\bf Variety}: this metric measures the variety of $N$ generated responses. Score 1 indicates that there is a significant difference in the linguistic patterns and wordings and score 0 otherwise.\\
{\bf Persona Detection}: we measure the model's ability to incorporate persona information by displaying two possible persona texts: one is the true persona that the model used and another is randomly sampled from the rest of persona texts. Then the judges will determine which is more likely to be the persona used by model according to $N$ generated responses. 

We calculated the Fleiss’ kappa to measure inter-rater consistency. Fleiss’ kappa for {\it Engagingness}, {\it Variety} and {\it Persona Detection} is 0.5149, 0.8281 and 0.6656, indicating ``moderate agreement'', ``almost perfect'' and ``substantial agreement'' respectively.

Results in Table ~\ref{tab:human_main} support: {\bf First}, Our model performs better than all baselines in all subjective metrics. {\bf Second}, Our model successfully incorporates persona information. {\bf Third}, Our model delivers diverse and engaging responses.

\begin{table}
\centering
\begin{tabular}{lccc}
\hline
{\bf Methods}&{\bf Engage.}&{\bf Variety}&{\bf Per. Detection}\\
\hline
Seq2Seq& 48.91\% & 26.07\% & 44.57\%\\
CVAE& 55.43\% & 88.04\% & 41.30\%\\
\hline
Per.-Seq2Seq& 64.13\% & 29.35\% & 69.57\%\\
GPMN& 67.39\% & 30.43\% & 65.76\%\\
Oracle. Copy& 63.26\% & 32.61\% & 72.83\% \\
\hline
Per.-CVAE& {\bf 71.74\%} & {\bf 90.22\%} & {\bf 78.26\%} \\
\hline
\end{tabular}
\caption{Human evaluation results ($N$=5). As a reference, the persona detection rate of the sampled testing data (human) is 90.83\%.}
\label{tab:human_main}
\end{table}

\begin{table}
\centering
\begin{tabular}{lccc}
\hline
{\bf Pref. (\%)} & Per.-Seq2Seq & GPMN & Per.-CVAE\\
\hline
Per.-Seq2Seq& - & 41.86 & 32.56 \\
GPMN& 58.14 & - & 39.53 \\
Per.-CVAE& {\bf 67.44} & {\bf 60.47} & - \\
\hline
\end{tabular}
\caption{Pairwise preference of three persona-based models ($N$=5).}
\label{tab:human_vs}
\end{table}

In addition, we performed a {\bf Preference Test} in three persona-based models. 
In this setting, the judges are presented with pairwise items and are asked to decide which of the two $N$ responses is of higher quality. And a partial ordering relation about the overall quality is given: best response quality $>$ response variety $>$ persona coverage. The judges first need to do a re-rank in mind and decide which model offers a better top response. If no difference then considering variety in responses, otherwise whether persona text is expressed.
As shown in Table ~\ref{tab:human_vs}, the proposed model is significantly (2-tailed t-test, $p < 0.01$) preferred than the other two baselines. In the pairwise setting, the diverse persona-based responses are more attractive to users. 

We present some generated examples in Table~\ref{tab:casestudy}.

\begin{table}
\centering
\begin{tabular}{c|l}
\hline
\multirow{3}*{ \rotatebox{90}{\it Mem}} &{\bf Chatbot's Persona:}\\
&{\quad \ \ I like tacos \quad\quad I have one brother}\\
&{\quad \ \ I love dogs \quad\quad I have three sisters}\\
\hline
\multirow{6}*{ \rotatebox{90}{\it 1st Turn}}&{\bf U:} Hi, how’s it going this afternoon ? \\
\cline{2-2}
&$r_1$: I have been doing pretty good . You?\\
&$r_2$: I am tired from working a day.\\
&$r_3$: Good. I just got done {\it eating some tacos}.\\
&$r_4$: I am watching Battlestar Galactica. How \\&\quad\ \ about you ?\\
\hline
\multirow{6}*{ \rotatebox{90}{\it 2nd Turn}} &{\bf U:}  Good. Tired from walking dogs early in the morning. \\
\cline{2-2}
&$r_1$: Wow! What are you into?\\
&$r_2$: I {\it love dogs} and I work in the fields.\\
&$r_3$: I am good at that. Do you have a job? \\
&$r_4$: Ok. I {\it like tacos}, and I am in Florida. Where \\&\quad\ \ are you from? \\
\hline
\end{tabular}
\caption{Generated examples of diverse responses in two consecutive turns. The italics show the incorporated persona.
$Mem$ denotes persona memory (in textual description) and $U$ denotes user input.
}
\label{tab:casestudy}
\end{table}

\subsection{Ablation Tests}

\begin{table}
\centering
\begin{tabular}{lccc}
\hline
{\bf Methods}&{\bf Dtinct-1}&{\bf Dtinct-2}&{\bf P. Cover}\\
\hline
Per.-CVAE& .0120 & {\bf .1037$^{\star}$} & {\bf .0410$^{\star}$} \\
w/o SDS& {\bf .0131} & .0921 & .0051 \\
w/o FDS& .0107 & .1021 & .0358\\
w/o SDS \& FDS& .0119 & .0907 & .0042 \\

\hline
\end{tabular}
\caption{The ablation tests ($N$=5). Results in {\it Dtinc-2} and {\it P.Cover} are significantly different from the ablated models (p-value$<$0.01).}
\label{tab:ablation}
\end{table}

In order to investigate the influence of different decoding strategies, we also conducted ablation tests where one or two decoding strategies were removed from Persona-CVAE. The results are shown in Table {\ref{tab:ablation}}. As we can see, after removing the SDS, the {\it Persona Coverage} decreases the most, indicating the SDS contributes to a higher persona information coverage. However, SDS also shows a negative effect on {\it Distinct-1}. This may be because the SDS encourages the use of words from profile sentences, which leads to low-frequency words more difficult to appear.

\section{Conclusion and Future Work}
In this paper, we focus on the diverse generation of conversational responses based on chatbot's persona and propose a memory-augmented architecture named Persona-CVAE. We conduct experiments on the ConvAI2 dataset. Experimental results show that our model outperforms the state-of-the-art methods, especially in terms of diversity and persona integration. For future work, we will explore modeling the persona information of users in open-domain conversations.

\section*{Acknowledgments}

The paper is supported by the National Natural Science Foundation of China under Grant No.61772153. We also thank the anonymous reviewers for their helpful comments.

\bibliographystyle{named}
\bibliography{ijcai19}

\end{document}